\documentclass{article}

\usepackage{arxiv}

\usepackage[utf8]{inputenc} 
\usepackage[T1]{fontenc}    
\usepackage{hyperref}       
\usepackage{url}            
\usepackage{booktabs}       
\usepackage{amsfonts}       
\usepackage{nicefrac}       
\usepackage{microtype}      
\usepackage{lipsum}

\usepackage{amsmath,graphicx}

\usepackage{hyperref}
\usepackage{cleveref}
\usepackage{multirow}
\usepackage{color}
\usepackage{diagbox}
\usepackage{xcolor,colortbl}
\usepackage{enumitem}
\usepackage{amssymb}
\usepackage{balance}

\newcommand{\ie}{i.\,e.\,, }

\newcommand{\et}{{et al.\,}}
\newcommand{\cf}{{cf.\,}}

\title{VOICE COMMAND GENERATION USING PROGRESSIVE WAVEGANS}

\author{
  Thomas Wiest\\
  ZD.B Chair of Embedded Intelligence\\ for Health Care and Wellbeing\\
  University of Augsburg, Germany\\
  \texttt{thomas.wiest@student.uni-augsburg.de} \\
  \And
  Nicholas Cummins\\
  ZD.B Chair of Embedded Intelligence\\ for Health Care and Wellbeing\\
  University of Augsburg, Germany\\
  \texttt{nicholas.cummins@ieee.org} \\
  \And
  Alice Baird\\
  ZD.B Chair of Embedded Intelligence\\ for Health Care and Wellbeing\\
  University of Augsburg, Germany\\
  \And
  Simone Hantke\\
  ZD.B Chair of Embedded Intelligence\\ for Health Care and Wellbeing\\
  University of Augsburg, Germany\\
  \And
  Judith Dineley\\
  ZD.B Chair of Embedded Intelligence\\ for Health Care and Wellbeing\\
  University of Augsburg, Germany\\
  \And
  Bj\"{o}rn Schuller \thanks{Bj\"{o}rn Schuller is also with Group on Language, Audio, and Music, Imperial College London, UK }\\
  ZD.B Chair of Embedded Intelligence\\ for Health Care and Wellbeing\\
  University of Augsburg, Germany\\
}

\begin{document}
\maketitle

\begin{abstract}
Generative Adversarial Networks (GANs) have become exceedingly popular in a wide range of data-driven research fields, due in part to their success in image generation. Their ability to generate new samples, often from only a small amount of input data, makes them an exciting research tool in areas with limited data resources. One less-explored application of GANs is the synthesis of speech and audio samples. Herein, we propose a set of extensions to the WaveGAN paradigm, a recently proposed approach for sound generation using GANs. The aim of these extensions -- preprocessing, Audio-to-Audio generation, skip connections and progressive structures -- is to improve the human likeness of synthetic speech samples. Scores from listening tests with 30 volunteers demonstrated a moderate improvement (Cohen’s $d$ coefficient of $0.65$) in human likeness using the proposed extensions compared to the original WaveGAN approach.
\end{abstract}


\section{Introduction}
\label{sec:intro}

Today, arguably one of the biggest applications of audio generation is text-to-speech synthesis. From public service announcements, through to chatbots and digital assistants, synthesised voices are becoming omnipresent in everyday life. Tech giants such as Amazon and Google are investing heavily in improving the authenticity -- the human likeness~\cite{Baird18-TPA} -- of these voices~\cite{Seppala2017, Robertson2016}. Google's WaveNet generative approach is a state-of-the-art example~\cite{wavenet}.

Generating natural speech is a highly non-trivial task. Predominant approaches in the literature over the years have included \emph{concatenative synthesis}~\cite{Hunt96-USI}, and \emph{statistical parametric speech synthesis}~\cite{Zen09-SPS}. Concatenative synthesis approaches are based on the concept of unit selection that stitches together small units of pre-recorded waveforms. Such an approach is comparatively simple when compared to other synthesis paradigms~\cite{Ling2015-DLF}. However, the generated speech can lack human likeness due to effects relating to boundary artefacts.

Statistical parametric speech synthesis approaches help avoid issues such as boundary effects by utilising acoustic models such as \emph{Hidden Markov Models} (HMMs) and/or \emph{Deep Neural Networks} (DNN)~\cite{Ling2015-DLF} to generate a smoothed trajectory of speech features for synthesis in a vocoder. However, due to effects such as oversmoothing~\cite{Zen09-SPS}, or inaccurate acoustic models~\cite{Zen09-SPS}, speech generated using such methods is often described as having a muffled and unnatural quality. At worst, the effects can result in generated speech that sounds even less human than that produced using concatenative synthesis~\cite{Zen09-SPS,Zen15-AMI}. Moreover, such systems tend to have highly complex processing pipelines that require extensive expertise and time to research and develop~\cite{Ling2015-DLF}. 

Newer approaches like Google's WaveNet~\cite{wavenet} have been developed to reduce the complexity of this pipeline by directly generating raw audio samples. WaveNet is a neural network based autoregressive approach which conditions each generated audio sample on a set of previously generated samples. WaveNet also handles longer-term temporal dependencies using a novel dilated causal convolutional structure~\cite{wavenet}.

Generative approaches based on adversarial networks have also begun to be explored for voice synthesis~\cite{Donahue18-SAW, Lee18-CWG}. These approaches are based on \emph{Generative Adversarial Networks} (GANs)~\cite{Goodfellow14-GANS} and use the associated zero-sum paradigm to directly generate new audio samples from a noise distribution. Despite the success of GANs in a range of applications~\cite{Creswell18-GAN}, speech samples synthesised using WaveGANs~\cite{Donahue18-SAW} are still clearly distinguishable from real human speech~\cite{Lee18-CWG}.

The work presented in this paper extends the original WaveGAN network. Specifically, we explore the benefits of encoder-decoder networks and progressive structures and adding simple processing methods. The autoencoder~\cite{pix2pix2016} and progressive~\cite{Wang18-TCG} inclusions allows us to condition the WaveGAN using other speech samples simplifying the overall learning task, while  the core goal of preprocessing is to ensure the maximum amount of speech-only samples are used in training. The presented results, gained using listening tests, indicate that our extensions improve the human-likeness of the generated samples when compared to the original WaveGAN structure.

\section{WAVEGANS PRELIMINARIES}
\label{sec:preliminaries}
In this section, the basic structures for understanding the WaveGAN approach are explained. This includes the basic GAN architecture (\cf \Cref{subsec:GAN}), the Wasserstein loss (\cf \Cref{subsec:Wasser}) function and the original WaveGAN network (\cf \Cref{subsec:WaveGAN}). 

\subsection{Generative Adversarial Networks}
\label{subsec:GAN}
The original GAN architecture proposed by Goodfellow \et \cite{Goodfellow14-GANS} is based on a two-player minimax game whose outcome is synthetically generated data samples. A GAN consists of a discriminator, $D$, and a generator, $G$. The generator's role is to learn to generate samples with a distribution $P_g$ from a target data distribution $P_x$ using an a priori known distribution $P_z$ to fool the discriminator into believing that instances sampled from the learnt distribution $P_g$ are from the actual real distribution. In most cases, the data sampled from $P_z$ are uniform random noise vectors with a value range from -1 to 1. The discriminator outputs single scalar values from 0 to 1 indicating the probability of a sample belonging to $P_g$ or $P_x$.

The generator and the discriminator are trained jointly in a minimax fashion formulated as:

\begin{equation}
V (G, D) = \mathbb{E}_{x \sim P_x}[log (D(x))] + \mathbb{E}_{z \sim P_z}[log (1 - D(G(z)))].
\end{equation}
\noindent
The associated value function $V (G, D)$ is minimised by the generator and maximised by the discriminator, \ie they try to optimise opposing objective functions in a zero-sum game. This set-up ensures that both networks compete against each other during training; the generator generates more and more realistic samples, while the discriminator becomes increasingly accurate at distinguishing the generated samples from the authentic data. For full details on the GAN training process, the interested reader is referred to~\cite{Goodfellow14-GANS}.

\subsection{Wasserstein GANs}
\label{subsec:Wasser}
Training instability is arguably one of the biggest limitations of GANs~\cite{gans_limits, improved-gan-training}. This can result in two major problems: \textit{mode collapse} and \textit{gradient vanishing}. In \textit{mode collapse}, the generator produces a very similar set of samples regardless of the input. In this case, the generator only learns a small subspace of the target data distribution. It switches to another small subspace as soon as the discriminator determines the current learnt model and labels all samples as generated; this process can continue indefinitely. \textit{Gradient vanishing} occurs when the discriminator achieves optimum performance early in training and is able to precisely differentiate between generated and real samples. Consequently, the generator cannot improve its data output any further as the required feedback from the generator is missing.

These issues are addressed by \emph{Wasserstein GANs}~\cite{wassersteingan}.
In this paradigm, the GAN learns a function that transforms the existing distribution $P_z$ into $P_g$ instead of directly learning the probability density function $P_g$ that approximates $P_x$ . This is achieved by calculating the \emph{Earth Mover} or \emph{Wasserstein} distance between $P_x$ and $P_g$. In this paper we use the regularised Wasserstein loss function:

\begin{equation}
\begin{split}
L = \mathbb{E}_{x \sim P_g}[D(x)] - \mathbb{E}_{y \sim P_r}[D(y)] \\ + \lambda \mathbb{E}_{m\sim P_m}[(||\nabla_m D(m)||_2 - 1)^2],
\end{split}
\end{equation}
\noindent
where $m$ is sampled from $x$ and $y$ and $t$ is uniformly sampled between 0 and 1.

\begin{equation}
    m = tx + (1-t)y
\end{equation}

This loss function ensures we avoid the negative effects of any weight clipping by including the gradient penalty which regulates the equation \cite{Improved-Wasserstein}. 

\begin{figure}[t]
	\centering
	\includegraphics[width=0.5\textwidth]{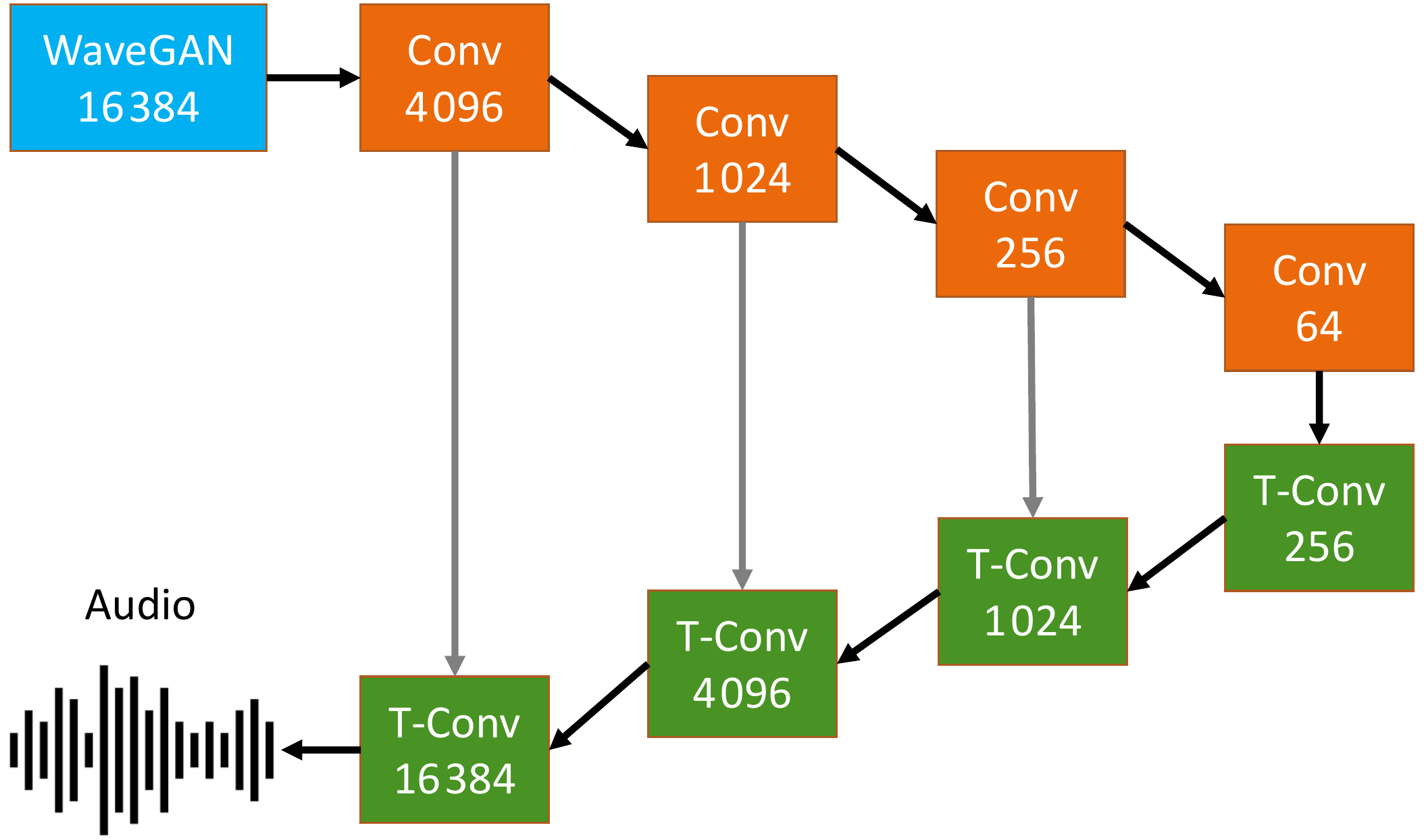}
	\caption{An illustrative example of our Audio-to-Audio generator architecture, enhanced with skip connections (in gray) using a WaveGAN as the input. Also shown are the size of the convolutional (Conv) and transposed convolutional (T-Conv) layers used during our experiments.}
	\label{fig:unet}
\end{figure}

\subsection{The WaveGAN Architecture}
\label{subsec:WaveGAN}
The original WaveGAN network, inspired by the \emph{deep convolutional GAN} (DCGAN) method proposed in~\cite{dcgan}, was the first attempt using GANs to synthesise raw audio~\cite{Donahue18-SAW}. It is similar to the DCGAN, but is naturally achieved using one-dimensional filters. The network uses a set of \emph{transposed convolution} operations to iteratively upsample and convert a 100-dimensional input vector (uniformly distributed noise values sampled between 0 and 1) into a high-resolution 16\,384-dimensional audio file. 

The discriminator essentially does the opposite process. It receives the audio files as an input, iteratively downsamples the input throughout its layers, and finally generates a single output value. The convolutional layers have the same kernel size and strides as the generator. Zero padding is used to avoid unwanted effects that otherwise result from shrinking the spatial dimensions. A \emph{phase shuffle} operation is also added between the convolutional layers of the discriminator, as the transposed convolution of the generator produces characteristic artefacts~\cite{odena2016deconvolution} when generating new samples. The operation randomly shifts individual vector entries by a small number to the left or the right. In doing so, it helps prevent the discriminator using the artefacts when distinguishing between the real and the fake samples.

\section{Progressive WaveGan Architecture}
In this section our proposed improvements to the WaveGAN architecture are described. This includes the preprocessing of audio sample in our datset (\cf \Cref{subsec:preproc}), the Audio-to-Audio generation extension (\cf \Cref{subsec:a2a}),  the inclusion of skip-connections (\cf \Cref{subsec:skip}), and the adoption of a progressive refinement structure (\cf \Cref{subsec:progstruc}).

\subsection{Preprocessing}
\label{subsec:preproc}
To improve training speed and data quality, we first do some simple preprocessing on the data. We observed that most samples in the \emph{speech command dataset} (\cite{speech-dataset}, \cf \Cref{subsec:data}) contain a small period of silence, both before and after the command. To avoid our GAN structure having to operate on this `silence', we estimate the actual beginning of the spoken word by calculating the short-term energy of the signal and applying a threshold. 

\subsection{Audio-to-Audio Generation}
\label{subsec:a2a}

To improve the quality of the audio output, we adopt a similar technique to the Image-to-Image Translation approach proposed in~\cite{pix2pix2016}. Thereby, instead of using a noise vector as the input to the generator, we take a \emph{conditional adversarial} approach and use other speech samples as the input. This approach should make the learning task more manageable as the input is already in a similar structure to the desired output.  We apply this technique in a \emph{command independent manner} on the \emph{speech command dataset}, \ie we do not condition the generator on the command we wish to generate. 

For this step, we adopt an autoencoder architecture (\cf \Cref{fig:unet}), with the input audio being compressed to smaller dimensions using stride convolutions. The output then represents a meaningful input vector which should still include some key structures that originate from speech. This representation then is used to generate new samples via transposed convolutions.

\subsection{Skip Connections}
\label{subsec:skip}
To further improve the stability of the autoencoder, we also introduced skip-connections~\cite{Mao16-IRU} between each convolutional and the corresponding transposed convolutional layer (\cf \Cref{fig:unet}). Skip connections pass data directly from one layer to another to skip the otherwise sequential data flow. These connections make it easier for the gradient to be back-propagated to bottom layers, thus helping to minimise gradient vanishing issues (\cf \Cref{subsec:Wasser}). Moreover, skip connections have been shown to aid speech enhancement paradigms~\cite{Park2017-AFC, Pascual2017-SSE}.

\subsection{Progressive Refinements Structure}
\label{subsec:progstruc}
\begin{figure}[t]
	\centering
	\includegraphics[width=0.5\textwidth]{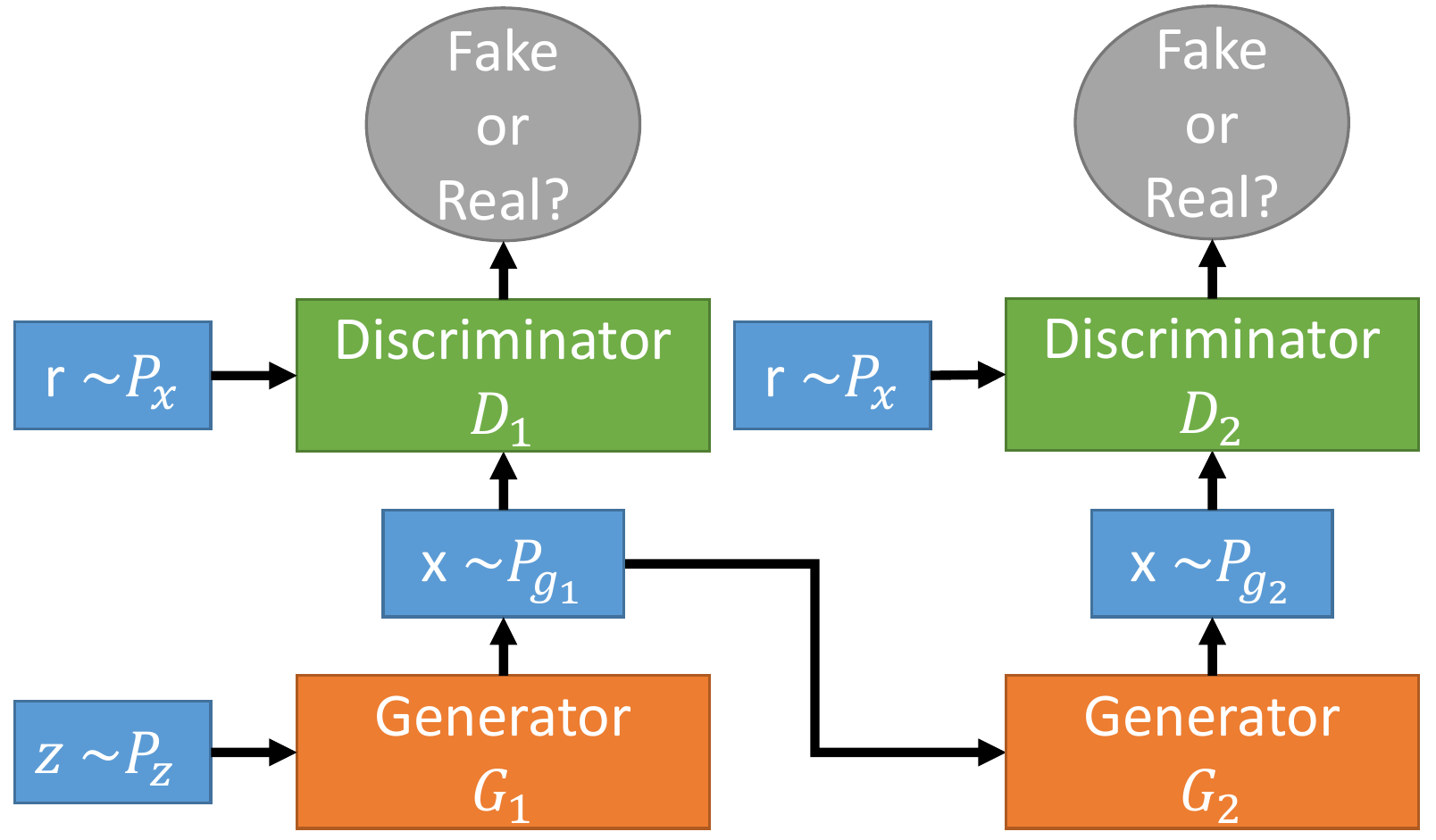}
	\caption{An example of a progressive GAN architecture in which the second GAN structure is conditioned on the generated samples from the first.}
	\label{fig:progressive}
\end{figure}

Even with the inclusion of preprocessing, Audio-to-Audio translation and skip connections, we cannot realistically expect our WaveGAN to output `perfect' speech. Therefore we adapt a progressive refinement scheme, as presented in~\cite{Wang18-TCG}, in which successive GAN structures are concatenated together to improve the overall speech generation quality (\cf \Cref{fig:progressive}). This extension was inspired by two main observations. Firstly, this approach has been shown to improve the quality of generated \emph{Positron Emission Tomography} (PET) image, and to the best of the authors' knowledge has not been applied for audio generation. Secondly, autoencoders are commonly used in a variety of audio application for denoising~\cite{vincent2010stacked, Weninger14-DRD}. 

\section{EXPERIMENTAL SETTINGS}
This section outlines our key experimental settings.

\subsection{Data}
\label{subsec:data}
For training the network, a subset of the Google speech command dataset is used \cite{speech-dataset}. Specifically, we use all spoken digits from zero to nine as an analogy to the widely used MNIST dataset. Our dataset therefore contained 23,\,666 samples distributed almost equally across the  ten commands. The total length of all commands 6:29:02 (hr:min:sec) at a mean length of 0.986s and a standard devation of 0.057s.

\subsection{WaveGAN Parameters}
We use a vector consisting of 100 random uniform noise values between 0 and 1 as input for the generator. The input first goes to a fully connected layer with a ReLU (Rectified Linear Unit) activation function. Now, instead of doing two dimensional transposed convolutions with a kernel size of 5x5 and strides of 2x2, we do one dimensional transposed convolutions with a kernel size of 25 and strides of 4. This is repeated until we reach the desired output size of (16\,384, 1). On a sample rate of 16\,000 Hz, this is slightly longer than a second, and should be enough to model spoken digits. The second dimension is the number of audio channels which will be 1 for all our experiments.

\subsection{Audio-to-Audio Parameters}
The Audio-to-Audio autoencoder architecture consists of four convolutions with a filter size of 25, a stride of four and a increasing amount of filters. The resulting data will be of size 64x512. This is scaled up again using four transposed convolutions with a decreasing amount of filters until we reach our goal dimension of 16384x1.

\subsection{Progressive Parameters}
In our experiments, we use two progress layers -- the first layer has to be a WaveGAN which takes a noise vector as input. The output of this network is then passed to an autoencoder to further improve the quality. We trained the first layer, the base architecture, for 140\,000 iterations and the second layer, the Audio-to-Audio architecture, for 20\,000 iterations.

\section{RESULTS AND DISCUSSION}
The evaluation of results of GANs is an open research topic with no universally agreed upon metric~\cite{Han18-ATI}. The widely used \emph{inception score}~\cite{barratt2018note} is not suitable for our evaluation for two reasons. Firstly, it was designed for evaluating images and thus requires the conversion of the wave files into spectrograms which is a lossy operation. Secondly, it does not necessarily correlate with the preferences of humans in case of audio files \cite{Donahue18-SAW}. We, therefore, use listening tests to evaluate our proposed architecture.

Using the \textsc{iHEARu-PLAY} annotation platform~\cite{Hantke15-IIA}, 30 participants evaluated a total of 20 audio samples; ten generated using the original WaveGAN and ten generated using our proposed approach\footnote{Samples available at: \url{https://goo.gl/sBGkxR}}. Each participant was asked to evaluate ``how human-like the audio samples sound'' on a 7 point Likert scale. In iHEARu-PLAY meta-data can be voluntarily given; $5$ out of our $30$ participants gave their details. The gender split was $4$ females, and $1$ male, the age range was $26-32$ years old with a mean age of $29$ years and a standard deviation of $\pm2.28$ years. 

The results of this evaluation revealed that listens found our samples from our proposed approach to be the more human-like (\cf \Cref{eval_table}). The samples produced by our proposed network achieved a mean rating of 4.48,  while those produced by the original WaveGAN network achieved a mean rating of 3.39. To quantify the observed effect we used the Cohen's $d$ measure, calculated to be $0.65$. As our $d$ coefficient is between $0.5$ and $0.8$ we can conclude this a medium effect~\cite{cohen1992power}.

\begin{table}
\caption{The results of our human listening evaluations. 30 participants listened to 10 files generated using the original WaveGAN network, and 10 files using our proposed approach, and ranked the human-likeness of each file on a 7 point Likert scale (1 not human, 7 human).}
\vspace{5pt}
\centering
\begin{tabular}{@{}lrrr@{}}
\toprule
Network & Mean Score & Std. Dev & Cohen's $d$ \\ \midrule
WaveGAN & $3.39$ & $\pm1.67$   & \multirow{2}{*}{$0.65$}  \\
Proposed Approach & \textbf{$4.48$}  & $\pm1.70$  &                               \\ \bottomrule
\end{tabular}
\label{eval_table}
\vspace{-5pt}
\end{table}

\subsection{Limitations of the Proposed Approach}
We observed during system development that both our network and the original WaveGAN approach were not always stable and could produce results of widely varying quality. More specific to our approach, the progressive structure meant our network has a longer training time than the original WaveGAN network. This increased training time limited the amount of hyperparameter optimisation we could realistically achieve. The conditional Audio-to-Audio extension also induced a new style of error. We produced incomprehensible words in approximately 5\,\% of cases by the mixing up two different commands. This effect was reduced, however, by the introduction of the skip connections. Finally, we faced a commonly occurring issue within GANs; aside from human evaluation, there is no appropriate objective evaluation metric to assess the quality of the generated samples.


\section{CONCLUSION AND FUTURE WORK}
\label{sec:conc}

Within this paper, we proposed a set of extensions to the WaveGAN approach for generating audio samples~\cite{Donahue18-SAW}. These extensions focused on improving the human-likeness of the generated samples. We first prepossessed the audio files to minimise the amount of silence feed into the network. We then modified the WaveGAN structure to be able to perform Audio-to-Audio generation via the inclusion of an autoencoder architecture with skip connections. Finally, we expanded this approach in a progressive structure to help refine the quality of the generated samples

Human evaluations indicated that our approach achieved the goal of improving the human-likeness over the original WaveGAN approach. However, there is still plenty of room for improvement in this regard. In particular, considerably more work is needed to help improve the stability of GANs regarding being able to consistently generating high-quality audio samples. In this regard, we will explore the benefits of using other loss functions to improve the stability of the adversarial training process.

\section{ACKNOWLEDGEMENTS}
This research  has received funding from the EU's 7$^{\mathrm{th}}$ Framework Programme ERC Starting Grant No.\ 338164 (iHEARu), the Bavarian State Ministry of Education, Science and the Arts in the framework of the Centre Digitisation.Bavaria (ZD.B).



\end{document}